\documentclass{article}

\PassOptionsToPackage{numbers, compress}{natbib}

\usepackage[preprint]{neurips_2024}

\usepackage[utf8]{inputenc} 
\usepackage[T1]{fontenc}    
\usepackage{hyperref}       
\usepackage{url}            
\usepackage{booktabs}       
\usepackage{amsfonts}       
\usepackage{nicefrac}       
\usepackage{microtype}      
\usepackage{xcolor}         

\usepackage{amsmath}
\usepackage{mathtools}
\usepackage{wrapfig}
\usepackage{algorithm}
\usepackage{algpseudocode}
\usepackage{multirow}
\usepackage{hyperref}
\usepackage{comment}

\title{Error Diffusion: Post Training Quantization with Block-Scaled Number Formats for Neural Networks}

\author{%
  Alireza Khodamoradi\thanks{Corresponding Author.}\\
  \and
  Kristof Denolf \\
  \and
  Eric Dellinger \\
  \AND
  \texttt{first.last@amd.com} \\
}

\begin{document}

\maketitle

\begin{abstract}
Quantization reduces the model's hardware costs, such as data movement, storage, and operations like multiply and addition. It also affects the model’s behavior by degrading the output quality. Therefore, there is a need for methods that preserve the model’s behavior when quantizing model parameters. More exotic numerical encodings, such as block-scaled number formats, have shown advantages for utilizing a fixed bit budget to encode model parameters. This paper presents error diffusion (ED), a hyperparameter-free method for post-training quantization with support for block-scaled data formats. Our approach does not rely on backpropagation or Hessian information. We describe how to improve the quantization process by viewing the neural model as a composite function and diffusing the quantization error in every layer. In addition, we introduce TensorCast, an open-source library based on PyTorch to emulate a variety of number formats, including the block-scaled ones, to aid the research in neural model quantization. We demonstrate the efficacy of our algorithm through rigorous testing on various architectures, including vision and large language models (LLMs), where it consistently delivers competitive results. Our experiments confirm that block-scaled data formats provide a robust choice for post-training quantization and could be used effectively to enhance the practical deployment of advanced neural networks.
\end{abstract}

\section{Introduction}\label{introduction}

The rise in popularity of neural models as an instrumental technology for integrating intelligence into electronic devices has increased the demand for innovative and efficient implementations of these models. They are constantly growing in size (number of parameters) and complexity (layer connections and their functions) \cite{9043731}. More parameters require more operations (e.g., multiply and addition) \cite{37631, canziani2017analysis}, more memory for storage \cite{8573527}, and larger bandwidth for data movement \cite{7783723}. Complex connections \cite{he2015deep, ronneberger2015unet} and complex functions (e.g., SoftMax) require caching, data movement orchestration, and more computation \cite{10.1145/3624990, 8631588}.

Quantizing model parameters enables the use of denser, faster, and cheaper operations \cite{10.1145/3316279.3316282}, and reduces memory, caching, and data movement requirements \cite{sohoni2022lowmemory,37631}. These substantial reductions in hardware costs are excellent incentives to quantize neural models and rigorously address and mitigate quantization’s negative effect on the model behavior and performance \cite{guo2018survey, gholami2021survey, li2023bivit}. 

The negative quantization effect on neural models is also known as the quantization error, and it increases when smaller number formats are used for quantization. Converting from 32-bit number formats (fp32) to 16-bit number formats (e.g., fp16 and bfloat16) can be done through simple casting while quantizing to 8-bit number formats (e.g., fp8 and int8) is a more sensitive process and may require per-channel, per-tensor, or per-kernel scales \cite{nagel2021white, kuzmin2024fp8, 8578558, NEURIPS2018_e82c4b19}. Quantizing to sub-8-bit number formats (e.g., int4) is more challenging and requires sophisticated methods \cite{zhang2023posttraining, frantar2023gptq}, mainly due to limited precision and dynamic range of these number formats \cite{xiao2024smoothquant, 8416865}.

Block-scaled numerical formats \cite{Kahan1971ASO, Scherl2011} have recently gained popularity in AI applications due to their numerical advantages, such as increased dynamic range over non-scaled data formats \cite{ rouhani2023microscaling, rouhani2023shared, NEURIPS2020_747e32ab}. 

Current post-training quantization methods are not designed with block-scaled number formats in mind and do not capture their specific characters, such as the dependency created by the shared scale between the values in a block. They also focus only on the layers being quantized, and their algorithms do not process layers that are not quantized. 

The ED method is the first PTQ algorithm supporting power-of-two block-scaled number formats and is designed to respect the data layout of these number formats. Because a block-scaled number format with a block size of one is technically a non-block-scaled number format, the ED algorithm also works and supports non-block-scaled number formats. In addition, this work introduces TensorCast, a PyTorch-based library to emulate a variety of data formats to assist researchers and engineers in their experiments in neural model quantization. We provide examples of how to use the TensorCast and quantize neural models with block-scaled data formats. Our innovations in the Error Diffusion PTQ method can be summarized as:
 
\begin{itemize}
    \item A novel weighted adjust-and-quantize method.
    \item Specialized support for block-scaled number formats.
    \item Introducing TensorCast, an open-source PyTorch library to support block-scaled data formats for quantization 
\end{itemize}

The rest of this article is organized as follows: section \ref{sec:ptq} reviews neural model quantization and related work. Section \ref{sec:blockscaled} covers the details of the block-scaled data formats. Section \ref{sec:ed} describes Error Diffusion in detail. Section \ref{sec:tensorcast} introduces TensorCast library. Section \ref{sec:results} discusses our results, and section \ref{sec:discussion} provides further discussion and our conclusions.

\section{Neural Model Quantization}\label{sec:ptq}

The neural network model is a set of layers, and its architecture connections are described with a directed acyclic graph (DAG). The model layers are the edges of this DAG, and they are functions mapping activation tensors (the nodes in the DAG) to activation tensors. 

\begin{wrapfigure}{r}{0.5\textwidth}
\begin{center}
    \includegraphics[width=0.5\textwidth]{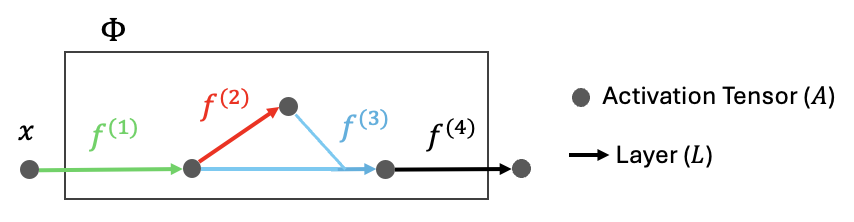}      
\end{center}
\caption{Model $\Phi=f^{(4)}\circ f^{(3)}\circ(f^{(2)}\circ f^{(1)}, f^{(1)})$ as a composite function with dependencies described by its DAG. Each function's connections is color-coded to show its inputs and output.}
\label{fig:simplephi}
\end{wrapfigure}

For example, model $\Phi$ in Figure \ref{fig:simplephi}, is composed of four layers. The input to $\Phi$, $x$ is the input to $f^{(1)}$ and its output, $f^{(1)}(x)$ is input to $f^{(2)}$ which maps it to $f^{(2)}(f^{(1)}(x))$. $f^{(3)}$ takes two inputs, one is the output of $f^{(1)}$ and the other one is the output of $f^{(2)}$ and can be shown as $f^{(3)}(f^{(2)}(f^{(1)}(x)), f^{(1)}(x))$. In this text, we adopt a more simple form by using the symbol $\circ$ and rewrite this relation as $f^{(3)}\circ (f^{(2)}\circ f^{(1)}, f^{(1)})$. Therefore, the model composite function composed of these four layer functions is shown as $\Phi=f^{(4)}\circ f^{(3)}\circ(f^{(2)}\circ f^{(1)}, f^{(1)})$.

Let's assume $f^{(2)}$ is the affine map of a linear layer $f^{(2)}(A)=AW^T+b$. Quantizing the weight matrix $W$ to $\widehat{W}$, changes this map to $\widehat{f}^{(2)}(A)=A\widehat{W}^T+b$ and, consequently, changes the model function to $\widehat{\Phi}=f^{(4)}\circ f^{(3)}\circ(\widehat{f}^{(2)}\circ f^{(1)}, f^{(1)})$.

One way to correct this change is to compile the new model function, $\widehat{\Phi}$, to behave similar to $\Phi$. The training process of a neural network is the compilation process of the model function. Applying quantization during the model training stage and correcting for its effect on the model function is known as quantization-aware training (QAT) \cite{park2018valueaware, qu2020adaptive, Shen_2021_ICCV, pmlr-v162-nagel22a, shen2019qbert}. 

In the QAT process, the effect of the quantization on the model’s output is captured by a loss function. The backpropagation can then suggest changing any model parameter to optimize for the loss function. Therefore, any combination of parameters in the whole model could be changed to correct the quantization effect. This process provides much flexibility and potential for compiling a quantized model with high performance. However, the training process requires vast computational resources \cite{10006884}, human expertise to adjust the training script \cite{NEURIPS2022_c1e2faff, hoffmann2022training}, and access to the training dataset that could be restricted by proprietary restrictions.

Another way to quantize a model is to apply the quantization after the training stage. This is also known as post-training quantization (PTQ). The PTQ process is computationally simpler than the QAT and does not require the training dataset. Given the high cost of training large language models \cite{NEURIPS2022_c1e2faff, musser2023cost}, PTQ is a cost-efficient option for quantizing and calibrating a neural model for a specific task(s).

However, quantizing trained parameters significantly and adversely changes the model’s behavior. Therefore, it is preferred for the PTQ methods to include methods to correct this change and reduce the quantization error. Popular PTQ algorithms measure the quantization changes in the model and suggest better rounding or update terms for model parameters \cite{xiao2024smoothquant, frantar2023gptq, zhang2023posttraining}. In this text, we refer to this process in PTQ, as model calibration.

The calibration of the model function in PTQ can be divided into multiple more straightforward tasks of calibrating its subfunctions. This is referred to as layer-wise calibration. Early works (\cite{nahshan2020loss, jacob2017quantization}) proposed the calibration in PTQ as minimizing the distance between the weights of each layer, $W^{(l)}$, and their quantized version, $\widehat{W}^{(l)}$, as:

\begin{equation}\label{eq:simple_calibration}
    \widehat{W}^{(l)} = \arg\min_{\widehat{W}^{(l)}}||W^{(l)}-\widehat{W}^{(l)}||_2^2
\end{equation}

Other works, such as \cite{nagel2020down, pmlr-v162-nagel22a}, proposed improving Equation \ref{eq:simple_calibration} by measuring the changes at the layer’s output instead of its parameters. Taking $\Phi$ in Figure \ref{fig:simplephi} as an example, quantization converts the weights, $W$, of each function, $f$, from a higher-precision number format to $\widehat{W}$ represented in simpler, lower-precision number format. Therefore, the output of layer function $l$ changes from $f^{(l)}(A,W^{(l)})$ to $f^{(l)}(\widehat{A},\widehat{W}^{(l)})$. Here, $\widehat{A}$ denote the quantization effect of every layer before layer $l$ in the model DAG on this layer’s input. A distance function $\mathcal{D}$ can measure the difference in each layer output to indicate functional change caused by the quantization. In this case, the calibration process in PTQ aims to minimize this distance instead of the one shown in Equation \ref{eq:simple_calibration}. 

A popular choice for $\mathcal{D}$ is the $l_2$-norm \cite{frantar2023optimal, frantar2023gptq, zhang2023posttraining}. Hence, the calibration goal is to find $\widehat{W}^{(l)}$ as:

\begin{equation}\label{eq:weight_change}
    \arg\min_{\widehat{W}} ||AW^{T} - \widehat{A}\widehat{W}^{T}||_2^2
\end{equation}

The OBS framework \cite{NIPS1989_6c9882bb, 298572} assumes $\widehat{A}=A$ and suggests quantizing each weight using the inverse Hessian information and updating the remaining non-quantized weights for PTQ calibration. However, in its proper form, this approach is computationally very demanding at $O(n^4)$ total runtime, with $n$ being the number of model parameters. To overcome this complexity, \cite{frantar2023optimal} sets $\widehat{A}=A$ in Equation \ref{eq:weight_change} and changes the $l_2$-norm term to $\sum_i||AW^T_{i,:}-A\widehat{W}^T_{i,:}||_2^2$. This way, all the $W$ rows have standard least squares form, and the Hessian calculation will be reduced to $H=2XX^T$. Later, GPTQ \cite{frantar2023gptq}, optimized this Hessian calculation by using the same order for all the rows and improving its numerical stability using Cholesky reformulation.

In this work, we also use the l2-norm to measure the changes in each output. Our method is based on GPFQ \cite{zhang2023posttraining} and propagates the quantization effect through the model DAG ($\hat{A}\neq A$ in Equation \ref{eq:weight_change}). This means that if $f^{(1)}$ in Figure \ref{fig:simplephi} is quantized and replaced by $\widehat{f}^{(1)}$, the input to $\widehat{f}^{(2)}$ will be different than the input to $f^{(2)}$, and our algorithm tries to minimize the difference between $f^{(2)}$ and $\widehat{f}^{(2)}$ outputs when their inputs are different. 

In addition, our PTQ method also calibrates layers that are not being quantized and supports block-scaled data formats and their data layout (more detail in Section \ref{sec:ed}). Before explaining the ED algorithm, we briefly review the block-scaled data formats in the next section.

\section{Block-scaled number formats}\label{sec:blockscaled}

\begin{wrapfigure}{r}{0.5\textwidth}
\begin{center}
    \includegraphics[width=0.3\textwidth]{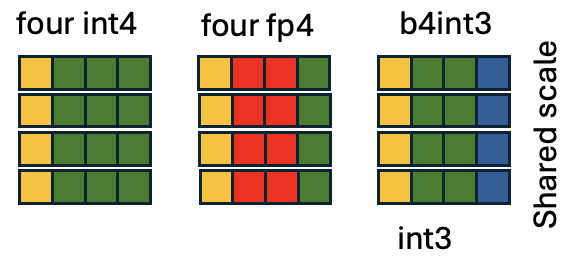}      
\end{center}
\caption{Bit layout for four values encoded in int4 (sign: 1 bit and mantissa: 3 bits), fp4 (sign: 1 bit, exponent: 2 bits, and mantissa: 1 bit), and b4int3 (sign: 1 bit, mantissa: 2 bits, and 4-bit scale).}
\label{fig:b4int3}
\end{wrapfigure}

A number format represents a range of numerical values within a budget of space measured in bits. It is a design choice that balances the number encoding capacity with energy and area. The resolution separating two close values is the number format’s precision, and its dynamic range indicates how far-apart values can be encoded by it. High precision and wide dynamic range require large bit budgets. The number of bits used by a number format indicates its alphabet size \footnote{alphabet size is the number of different possible encodings ($2^b$ for $b$ bits) that can be used to encode numbers, NaN, infinity, etc.}. However, number formats with larger bit budgets are more expensive to support in hardware. For example, the IEEE 754 fp32 number format can encode positive values from $1.4\times10^{-45}$ to $3.4\times10^{38}$ and has an alphabet size of $2^{32}$, while the fp8(e4m3) \cite{ocp_mx} number format can encode positive values from $1.9\times10^{-3}$  to $448$ and has an alphabet size of $2^8$. The storage and data movement for fp32 values requires four times more memory and energy than fp8, and modern computing units can perform 16x more floating operations on fp8 values than fp32 values in the same number of clock cycles \cite{mi300}.

Previous works \cite{rouhani2023microscaling, rouhani2023shared, NEURIPS2020_747e32ab} have shown the benefit of quantizing neural models with block-scaled number formats. These data formats have also gained large support from the computing hardware community \cite{ocp_mx}. Block-scaled data formats share a few bits between a group of values to scale them together. Proper scaling can help with encoding weights at the quantization time by better matching the weights’ distribution. It also creates a dependency between the values inside each block. 

For example, we can encode four values with int4, fp4, or b4int3 (Figure \ref{fig:b4int3}).  In the block-scaled number format, b4int3, each value has a private int3 data format, and every four values share a 4-bit scale. Int4 and fp4 encode $\pm\{0, 1, 2, 3, 4, 5, 6, 7\}$ and $\pm\{0, 0.5, 1, 1.5, 2, 3, 4, 6\}$, their dynamic ranges \footnote{in this text we define the dynamic range as largest-in-magnitude value divided by smallest-in-magnitude and non-zero value} are 7 and 12, and their precisions are 1 and 0.5, respectively. The alphabet size for int4 and fp4 is 15 ($+0=-0$). With b4int3, 16 different shared scales can scale the seven int3 values and create 67 unique numbers between $\pm3\times2^8$ and $\pm2^{-7}$. If each value's private bits encode a numerical value $p_i$ and shared scale bits encode a numerical value $s_b$ in the block-scaled number format, each value can be calculated as:

\begin{equation}\label{eq:blockscaled}
    w_i = s_b \times p_i
\end{equation}

The shared scale in b4int3 encodes a power of two number. Therefore the above equation is just a shift in hardware and is highly desired for low-power high-performance chips. Note that int4 and fp4 cannot encode anything smaller than $\pm0.5$ or larger than $\pm7$. While b4int3 can scale the int3 values to larger or smaller numbers. However, only four different numbers are available for encoding a block with each scale. Therefore, the quantization errors for all the values inside a block depend on the selected scale. This is the dependency inside a block encoded with block-scaled data format. The following section describes how our PTQ algorithm accounts for this dependency.

\section{Error Diffusion (ED)}\label{sec:ed}

\begin{figure}
\begin{center}
    \includegraphics[width=\textwidth]{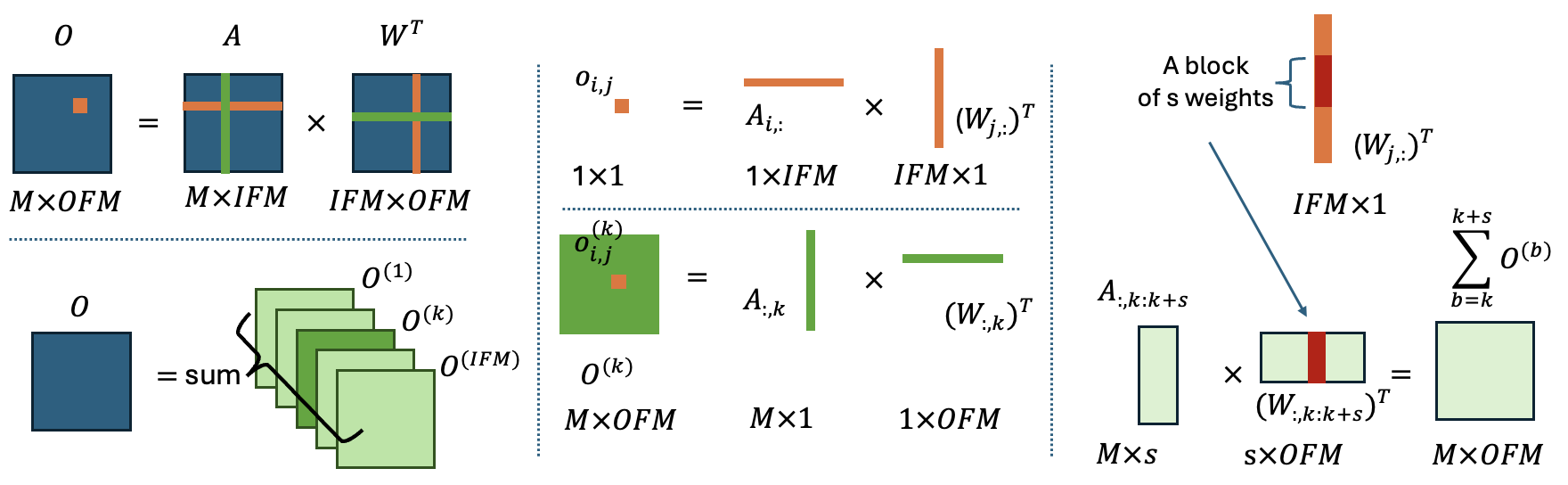}      
\end{center}
\caption{Top left: output, $O_{M\times OFM}$, generated by the matrix multiply of the input activations and weights. Top middle: each element of the output $o_{i,j}$ is the sum of products of two vectors with size $1\times IFM$. Bottom middle: $k$th portion of the output matrix is the outer product of two vectors with sizes $M\times 1$ and $1\times OFM$. Left bottom: output is the sum of its $IFM$ portions. Top Right: a block of $s$ numbers in $(W_{j,:})^T$ column. Bottom right: same block of $s$ numbers expands over $s$ rows $(W_{:,k:k+s})^T$ and effects $s$ output portions.}
\label{fig:dims}
\end{figure}

The output of a linear layer with input activations $A_{M\times IFM}$ and weight matrix $W_{OFM\times IFM}$, is $O_{M\times OFM} = AW^T$ (Figure \ref{fig:dims}, top left). We re-write this matrix multiply as following:

\begin{equation}\label{eq:o_k}
    O = \bigcup_{i=1}^M\bigcup_{j=1}^{OFM}\sum_{k=1}^{IFM} o_{i,j}^{(k)} = \sum_{k=1}^{IFM}\bigcup_{i=1}^M\bigcup_{j=1}^{OFM} o_{i,j}^{(k)} = \sum_{k=1}^{IFM} A_{:,k}(W_{:,k})^T = \sum_{k=1}^{IFM}O^{(k)} 
\end{equation}

Here, $o_{i,j} = \sum_{k=1}^{IFM} a_{i,k}*w_{j,k}$ is the output at $(i,j)$ (Figure \ref{fig:dims}, top middle), and $o_{i,j}^{(k)} = a_{i,k}*w_{j,k}$ is the $k$th portion of it (Figure \ref{fig:dims}, the orange square at bottom middle). $O^{(k)}=A_{:,k}(W_{:,k})^T$ (Figure \ref{fig:dims}, bottom middle) is the $k$th portion of $O$ (Figure \ref{fig:dims}, bottom left), and is generated by the $k$th column of $A$ and $W$ (or $k$th row of $W^T$).

Let's denote the quantized weights as $\widehat{W}$, and use $\widehat{A}$ to denote the activation tensors in the quantized model. For simplicity, we use $\widehat{W}_k^T$ instead of $(\widehat{W}_k)^T$, and $k$ instead of $(:,k)$. Therefore, $k$th portion of a quantized layer's output is $\widehat{O}^{(k)} = \widehat{A}_k\widehat{W}_k^T$.

We define the quantization error matrix as difference between the original and quantized outputs:

\begin{equation}\label{eq:errormatrix}
        E^{(k)} = A_kW_k^T-\widehat{A}_k\widehat{W}_k^T
\end{equation}

\cite{zhang2023posttraining} shows that the output of a convolution layer can also be analysed this way. 

\subsection{Weighted Error Matrix}

The greedy path-finding quantization (GPFQ) explained in \cite{zhang2023posttraining} is only aware of the accumulated error generated by processing the prior iterations (the term $U^{(k-1)}$). We propose to diffuse the quantization error by looking at the whole input matrices $A$ and $\widehat{A}$ at each $k$th step. Therefore, we start with the following goal:

\begin{equation}\label{eq:ed_goal}
    \sum_{k=1}^{IFM} E^{(k)} = (A-\widehat{A})W^T+\sum_{k=1}^{IFM} \widehat{A}_k(W^T_k-\widehat{W}^T_k) = \Tilde{O}+\sum_{k=1}^{IFM} \widehat{A}_k(W^T_k-\widehat{W}^T_k) = 0
\end{equation}

Here, $\Tilde{O}=(A-\widehat{A})W^T$ is the error in layer's input, inherited from prior quantized layers. Ideally, we need scalars $\alpha^{(k)}$ such that $\sum_{k=1}^{IFM}\widehat{A}_k(W^T_k-\widehat{W}^T_k)+\alpha^{(k)}\Tilde{O}=0$. Heuristically, We observed that $\alpha^{(k)}=\frac{1}{IFM}$ is a sufficient term to divide Equation \ref{eq:ed_zero} into the iterative following sub-goals:

\begin{equation}\label{eq:ed_zero}
    U^{(k)} = \frac{\Tilde{O}}{IFM} + \widehat{A}_k(W^T_k-\widehat{W}^T_k) + U^{(k-1)}= 0
\end{equation}

Therefore, ED's update and quantization equations are:

\begin{equation}\label{eq:ed_update}
    \widehat{W}^T_k = W^T_k + \frac{\widehat{A}^T_k}{||\widehat{A}_k||_2^2}(\frac{\Tilde{O}}{IFM} + U^{(k-1)})
\end{equation}

\begin{equation}\label{eq:ed_quant}
    \widehat{W}^T_k = \text{quant}\{W^T_k + \frac{\widehat{A}^T_k}{||\widehat{A}_k||_2^2}(\frac{\Tilde{O}}{IFM} + U^{(k-1)})\}
\end{equation}

Note that Equation \ref{eq:ed_update} has no hyperparameters and for $A=\widehat{A}$, it describes the same function as \cite{zhang2023posttraining}.

\subsection{Calibrating the neural model as a composite function}

In many quantization applications, some layers, such as the last layer, are kept unquantized \cite{zhang2023posttraining, frantar2023gptq}. Similarly, we also update these layers with Equation \ref{eq:ed_update} to better diffuse the error in the model. Take the model in Figure \ref{fig:simplephi}, let's assume we quantize layers $f^{(1)}$, $f^{(2)}$, and $f^{(3)}$ to $\widehat{f}^{(1)}$, $\widehat{f}^{(2)}$, and $\widehat{f}^{(3)}$ respectively and keep $f^{(4)}$ unquantized. Hence the quantized model is $\widehat{\Phi}=f^{(4)}\circ \widehat{f}^{(3)}\circ(\widehat{f}^{(2)}\circ \widehat{f}^{(1)}, \widehat{f}^{(1)})$.

Intuitively, all activation tensors are tensor spaces that encode information about the model input and are the same mathematical object. The quantization error flowing in the model through activation functions ($A-\widehat{A}$) are also the same mathematical objects as the activation functions. Layer functions map their input activation tensors to output activation tensors, in fact, they map mathematical objects to the same objects (endofunctions). This means the same mathematical objects are flowing through the neural model and they just change their shape and size from one layer to another. If, in a quantization application, only a subset of these functions (model layers) is to be quantized, the quantization error will also flow through the non-quantized layers. Therefore, the model behavior will benefit more from calibrating every layer to reduce the quantization error than just calibrating the subset of layers that are being quantized. In this case, Equation \ref{eq:ed_quant} will be used to calibrate and quantize the layers that are being quantized, and Equation \ref{eq:ed_update} will be used for non-quantized layers to just calibrate them.

\subsection{Support for block-scaled data formats}

Computing Equation \ref{eq:o_k} in hardware loops over the $IFM$ dimension to calculate $o_{i,j}=\sum_{k=1}^{IFM}a_{i,k}*w_{j,k}$ (Figure \ref{fig:dims}, top middle). In a hardware realization of this operation, it is preferred to group the weights in the same direction for block-scaled number formats (Figure \ref{fig:dims}, top right). Using the notation in Equation \ref{eq:blockscaled}, this loop is calculated as:

\begin{equation}\label{eq:block_loop}
    \sum_{k=1}^{IFM}a_{i,k}*w_{j,k} = \sum_{b=1}^{IFM/block\_size}\{s_b\times\sum_{k=1+(b-1)*block\_size}^{b\times block\_size}a_{i,k}*v_{j,k}\}
\end{equation}

This also means that each $v_{j,k}$ shares a block scale with a few other values in the $k$ dimension and each block expands over multiple error matrices shown in Equation \ref{eq:errormatrix}. Figure \ref{fig:dims}, right, displays the orthogonality of block direction and the direction of quantization correction.

And the block error matrix can be defined as $E_{block}^{(b)} = O_{block}^{(b)}-\widehat{O}_{block}^{(b)}$, and we use it to re-write our goal in Equation \ref{eq:ed_goal}:

\begin{equation}\label{eq:ed_goal_block}
    \Tilde{O}+\sum_{b=1}^{IFM/block\_size}\sum_{k=1+(b-1)*block\_size}^{b\times block\_size} \widehat{A}_k(W^T_k-\widehat{W}^T_k) = 0 
\end{equation}

Therefore, the update matrix for the block is:

\begin{equation}\label{eq:ed_zero_block}
    U_{block}^{(b)} = \frac{\Tilde{O}}{IFM/block\_size} + E_{block}^{(b)} + U_{block}^{(b-1)}
\end{equation}

All the values inside the block will be scaled with a shared scale. When processing step $k$, quantizing $W_k$ may effect the scale calculation and could change the previously quantized values in the same block. At each step $(1+(b-1)*block\_size\leq k \leq b*block\_size)$, the block error $E_{block}^{(b)}$ should be re-calculated. This means $U_{block}^{(b)}$ is updated at each iteration inside its block, $b$. In this case, at each iteration, $l$, inside the block, the following calculation is required:

\begin{equation}\label{eq:ed_l_update}
    l\_update=\frac{\Tilde{O}}{IFM/block\_size} + \sum_{k=1+(b-1)*block\_size, k\neq l}^{b\times block\_size} \widehat{A}_k(W^T_k-\widehat{W}^T_k) + U_{block}^{(b-1)}
\end{equation}

Note that the above sum  skips over $k=l$. This is the total update term for the block minus step $l$. By evenly distributing this error among $block\_size$ steps, at iteration $l$ the update term is:

\begin{equation}\label{eq:ed_block_update}
    \widehat{W}_l^T=W_l^T+\frac{\widehat{A}_l^T}{||\widehat{A}_l||_2^2}\frac{l\_update}{block\_size}
\end{equation}

Note that for $block\_size=1$, Equations \ref{eq:ed_update} and \ref{eq:ed_block_update} describe the same functions.

\subsection{The Issue with the Large M}

Similar to \cite{zhang2023posttraining, frantar2023gptq}, we also use multiple sample inputs for calibration. As explained in \cite{zhang2023posttraining}, the number of input samples linearly increases the size of $M$ as it is equal to $input\_size*number\_of\_samples$. This results in large memory requirement to store the error and update matrices. Equation \ref{eq:ed_l_update}'s three matrices, $\Tilde{O}$, $U_{block}^{(b-1)}$, and $\widehat{A}_k(W^T_k-\widehat{W}^T_k)$ are $M\times OFM$. To process each block, we can store the sum of the first two matrices as one entity and calculate the sum as following to minimize memory:

\begin{equation}\label{eq:ed_low_memory}
    \sum_{k=1+(b-1)*block\_size, k\neq l}^{b\times block\_size} \frac{\widehat{A}_l^T\widehat{A}_k}{||\widehat{A}_l||_2^2}(W^T_k-\widehat{W}^T_k)
\end{equation}

The term $\frac{\widehat{A}_l^T\widehat{A}_k}{||\widehat{A}_l||_2^2}$ is a scalar and the above sum requires $1\times OFM$ to store its results instead of $M\times OFM$. Additionally, one can pre-calculate all the $\frac{\widehat{A}_l}{||\widehat{A}_l||_2^2}(\Tilde{O}/(IFM/block\_size)+U_{block}^{(b-1)})$ and store a $block\_size\times OFM$ matrix on chip instead of an $M\times OFM$ one. For example, if we use $256$ samples for calibrating a fully connected layer in OPT \cite{zhang2022opt} with input size $2048$ and $OFM$ size $8192$, storing all three matrices in Equation \ref{eq:ed_l_update} requires $3\times2^{34}$ Bytes, while with pre-calculating and using Equation \ref{eq:ed_low_memory} for $block\_size=32$, we only need $2^{20}$ Bytes.

\section{TensorCast}\label{sec:tensorcast}
TensorCast is an open-source casting/quantization library based on PyTorch 2.2+. Its scope is defining number formats and converting tensors between them. It supports a variety of data formats, including the OCP MX datatypes described in \cite{ocp_mx} as well as additional data formats as needed to support research in the area of low precision for machine learning. TensorCast provides the needed infrastructure for researchers to describe new data formats and experiment with them to quantize neural models. The source code is available at \url{github.com/ROCm/tensorcast}.

\section{Results}\label{sec:results}

The ED algorithm simply selects the shared scale based on the largest-in-magnitude value in the block. Although improvements in better selecting the shared scale are part of our future work, our result section demonstrates that even with this intial approach, the results generated with block-scaled data formats are competitive with the tensor-scaled and row-scaled results achieved by other algorithms using a more compute costly tensor mean (GPFQ \cite{zhang2023posttraining} with hyperparameters) or Hessian information (GPTQ \cite{frantar2023gptq}) to adjust their scale.

To our knowledge, ED is the first published PTQ algorithm that uses block-scaled data formats described in this text. We provide Table \ref{tab:vision} to compare ED’s algorithm with row-scaled methods. In this table, our results on image classification models show a clear advantage for ED over other methods. Some of the methods in Table \ref{tab:vision} use a shared minimum in addition to the shared scale, which we do not use in the ED algorithm.

\begin{table}[ht]
\centering
\resizebox{0.9\textwidth}{!}{
\begin{tabular}{|c|c|c|c|c|c|c|c|}
\hline
Model                     & Bits & ED & GPFQ\cite{zhang2023posttraining} & MSE\cite{banner2019posttraining} & BRECQ\cite{li2021brecq} & AdaRound\cite{nagel2020down} & OMSE\cite{choukroun2019lowbit} \\ \hline
\multirow{2}{*}{ResNet18} & 3    &  \textbf{0.9679}    & 0.9539   &  -   &  -     &    -      &   -   \\ \cline{2-8} 
                          & 4    &  \textbf{0.9940}    & 0.9826   &  0.9612   &   \textbf{0.9940}    &    0.9860      &   0.9819   \\ \hline
\multirow{2}{*}{ResNet50} & 3    &  0.9394    &  \textbf{0.9431}  &  -   &   -    &     -     &   -   \\ \cline{2-8} 
                          & 4    &  \textbf{0.9910}    &  0.9864  &  0.9697   &   0.9907    &  0.9889        &  0.9655    \\ \hline
\end{tabular}}
\caption{Non block scaled data formats. Weights are quantized to int3 and int4 data formats. Results are normalized Top-1 accuracy (top-1 quantized divided by Top-1 baseline).}
\label{tab:vision}
\end{table}

To compare our results with the SOTA PTQ methods used for LLMs, we compare them with the method explained in \cite{frantar2023gptq}. Table \ref{tab:edvsgptq} shows our results with MX data formats \cite{ocp_mx} with per-block power-of-two shared scales against the Hessian-based algorithm using a per-row floating scale. The power-of-two scaling is cheaper in hardware implementation, and Table \ref{tab:edvsgptq} shows competitive results with the MX data formats. In this table, the \textit{ED+Calib} row includes the results of calibrating the last fully connected layer, as explained in subsection \ref{sec:ed}.2.

\begin{table}[ht]
\centering
\resizebox{0.6\textwidth}{!}{
\begin{tabular}{|c|c|c|c|c|c|c|}
\hline
OPT      & Bits               & 125M  & 350M  & 1.3B  & 2.7B  & 6.7B  \\ \hline
Baseline & 16                 & 27.56 & 22.00 & 14.63 & 12.47 & 10.86 \\ \hline \hline
GPTQ     & int4-fp & 31.12 & 24.24 & 15.47 & \textbf{12.87} & 11.39  \\ \hline 
ED       &   mxint4                 & 30.33  &   23.58    &  15.13     &   12.91    &  11.27    \\ \hline
ED+Calib       &   mxint4                 & \textbf{30.28}  &  \textbf{23.59}     &   \textbf{15.12}    &  12.91   & \textbf{11.29} \\ \hline \hline
GPTQ     & int3-fp & 53.85 & 33.79 & \textbf{20.97} & \textbf{16.88} & \textbf{14.86}  \\ \hline 
ED       &      mxint3              & 49.80 &  33.59     &   23.24    &   17.70    &  19.68  \\ \hline
ED+Calib       &   mxint3                 & \textbf{49.33}  &  \textbf{33.29}     &  22.03     &   17.34    &   18.94   \\ \hline
\end{tabular}}
\caption{Comparing float scaled per row \cite {frantar2023gptq} against power-of-two scaled MX data format \cite{ocp_mx} used in the ED algorithm on WikiText2.}
\label{tab:edvsgptq}
\end{table}

As explained in section \ref{sec:ed}, we account for the change in activations after quantization ($A\neq \widehat{A}$). With no modification in the algorithm, we apply quantization to activation and calibrate the weights. Table \ref{tab:res_both} displays our results on different models and datasets with activation and weight quantization with different four and six bit mini-float MX data formats.

\begin{table}[ht]
\resizebox{\textwidth}{!}{
\begin{tabular}{|c|c|c|c|c|c|cc|c|}
\hline
\multirow{2}{*}{Task} & \multirow{2}{*}{Family} & \multirow{2}{*}{Model} & \multirow{2}{*}{Dataset} & \multirow{2}{*}{Metric} & \multirow{2}{*}{FP32} & \multicolumn{2}{c|}{MXFP6} & \multirow{2}{*}{MXFP4} \\ 
\cline{7-8}
                    &                     &                    &                     &                     &                    & E2M3                  & E3M2                      &                     \\ \hline
                     & Vision          & DeiT-Tiny    &                          &                       & 72.16                 & \multicolumn{1}{c|}{72.16}  & 71.29  &  64.76     \\ \cline{3-3} \cline{6-9} 
       Image        &    Transformer  & DeiT-Small           &      ImageNet   &     Top-1         & 80.54                 & \multicolumn{1}{c|}{80.50}  & 80.25  & 76.80        \\ \cline{2-3} \cline{6-9} 
\multirow{2}{*}{Classification} & \multirow{3}{*}{CNN} & ResNet-18                 & \multirow{2}{*}{ILSVRC12} & \multirow{2}{*}{Acc. $\uparrow$} & 70.79  & \multicolumn{1}{c|}{70.66}  & 70.15  &  67.40                   \\ \cline{3-3} \cline{6-9} 
                    &                     & ResNet-19                 &                     &                     & 77.40                 & \multicolumn{1}{c|}{77.15}  & 76.48  & 69.99  \\ \cline{3-3} \cline{6-9} 
                    &                     & MobileNet v2          &                     &                     & 72.14                 & \multicolumn{1}{c|}{70.22}  & 65.32  & 18.88 \\ \hline
Speech              & \multirow{2}{*}{Transformer}  & \multirow{2}{*}{Wav2Vec 2.0}  & \multirow{2}{*}{LibriSpeech}  & \multirow{2}{*}{WER $\downarrow$} &\multirow{2}{*}{18.90}& \multicolumn{1}{c|}{\multirow{2}{*}{19.09}}  & \multirow{2}{*}{19.6}  & \multirow{2}{*}{24.39}                    \\
Recognition         &                               &                               &                                &                     &                    & \multicolumn{1}{l|}{}  &   &                     \\ \hline
\end{tabular}}
\caption{Both the activations and pre-trained weights
from the baseline model are quantized to the column’s datatype.}
\label{tab:res_both}
\end{table}

\section{Conclusions}\label{sec:discussion}
This paper reviewed the block-scaled data formats and introduced the ED PTQ method. We explained the novel and hyperparameter-free structure of the ED algorithm for supporting these number formats and why calibrating all layers, including the ones kept in full precision improves the PTQ results. By having a shared scale, the block-scaled data formats introduce a dependency between the values inside the block.  Therefore, selecting a good scale is a crucial step in the quantization process. Our results show that the low cost way of calculating the shared scale based on the largest absolute value in the block, these data formats can provide competitive results with a proper PTQ method. As for our future work, we plan to find ways to improve the shared scale calculation for PTQ of neural models.

\section{Limitations}
The ED algorithm introduced in this paper is the first PTQ method that uses the power of two block-scaled data formats. Shown results are comparable to float-scaled data formats. However, ED is not tested in cases where higher precision scales are needed (e.g., the presence of significant outliers in data).
The sample size of the calibration data is fixed for each model class (e.g., OPT, ResNet) and is heuristically set to match the available GPU memory.
The block-scaled data formats introduced in \cite{ocp_mx} are subject to change in future revisions. Although ED is not tested against future data formats, we believe our method should hold its performance. 

\bibliographystyle{unsrtnat}
\bibliography{main.bib}
\end{document}